\documentclass{article}
\usepackage{ijcai15}
\usepackage{graphicx}
\usepackage{algorithm,algorithmic}
\usepackage{times}

\title{Exploring Bayesian Models for Multi-level Clustering of Hierarchically Grouped Sequential Data}
\author{Adway Mitra \\
Indian Institute of Science\\
Bangalore, India \\
adway@csa.iisc.ernet.in}

\begin{document}

\maketitle

\begin{abstract}
A wide range of Bayesian models have been proposed for data that is divided hierarchically into groups. These models aim to cluster the data at different 
levels of grouping, by assigning a mixture component to each datapoint, and a mixture distribution to each group. Multi-level clustering is 
facilitated by the sharing of these components and distributions by the groups. In this paper, we introduce the concept of Degree of Sharing (DoS) for the mixture components and 
distributions, with an aim to analyze and classify various existing models. Next we introduce a generalized hierarchical Bayesian model, of which the existing models can be shown 
to be special cases. Unlike most of these models, our model takes into account the sequential nature of the data, and various other temporal structures at different levels while 
assigning mixture components and distributions. We show one specialization of this model aimed at hierarchical segmentation of news transcripts, and present a Gibbs Sampling 
based inference algorithm for it. We also show experimentally that the proposed model outperforms existing models for the same task.
\end{abstract}

\section{Introduction}
In many applications we come across hierarchically grouped data. For example in a text corpus, data is grouped into documents, paragraphs and sentences. Such data can be clustered
 at multiple levels, based on the notion of topics. A large number of hierarchical Bayesian models have been proposed for such data, many of whom are quite similar to each other 
in various aspects. However, to the best of our knowledge, there has not been much research aimed at placing these models in perspective, and making a comparative study of them, 
except empirical comparisons. This is what we attempt in this paper. The main aspect of these models which we compare is how they share the mixture components and distributions 
across the groups at different levels.

The contributions of this paper are as follows: 1) We introduce a novel classification of Hierarchical Bayesian models for grouped data, based on Degree of Sharing of mixture 
components and distributions 2) We introduce a generalized Hierarchical Bayesian model and show many existing ones to be special cases of it, and 3) We show how it can be 
adapted for news transcript segmentation, for which we give an inference algorithm and demonstrate experimental results.
%

\section{Notations}
\begin{figure}
 \centering
\includegraphics[width=3.5in,height=1.4in]{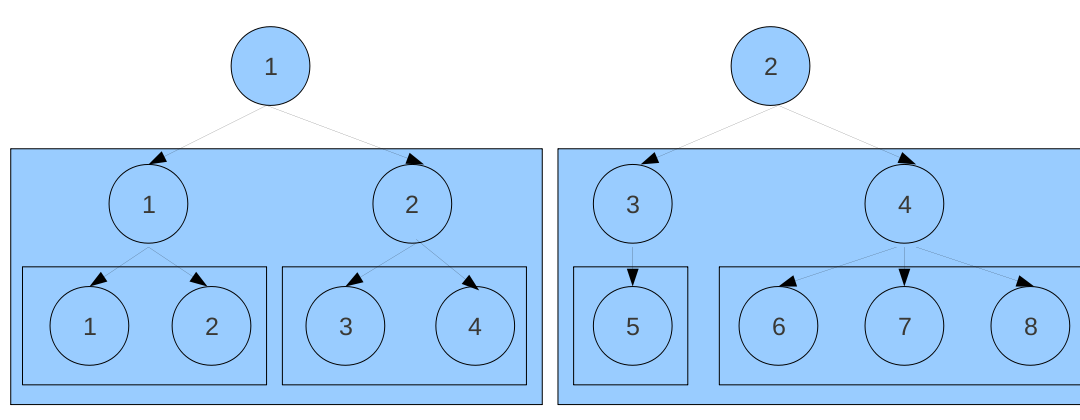} 
\caption{\small{Data grouped at 3 levels: $D^1_i=i \forall i$, $D^2_1=1$,$D^2_3=2$,$D^2_5=3$,$D^2_6=4$ etc, $D^3_i=1$ for $i=1\dots 4$, $D^3_i=2$ for $i=5\dots 8$. 
Also, $D^3(1)=1$, $D^3(3)=2$ etc}}
\label{fig:group}
\end{figure}

Consider $N$ datapoints $Y_1,Y_2,\dots,Y_N$, of any type (eg. integers, real-valued vectors) based on the application.  Each 
of these are associated with \emph{group membership variables} (positive integers), which specify the grouping of the datapoints. If there are $L$ 
levels of grouping, each datapoint $Y_i$ is associated with observed variables $\{D^1_i,D^2_i,\dots,D^L_i\}$. For example, a text corpus consists of a set of documents, each of 
which consists of word-tokens. We can consider the word-tokens as data-points $\{Y_i\}$, which are tagged with their document memberships using $\{D^2\}$, where $\{D^1\}$ are the 
token indices, to capture the sequential ordering. This is the standard setting used in most topic models for text documents. In addition, it is possible to consider a 3-level 
grouping with \emph{sentences} within documents. Then each word-token $Y_i$ is associated with a sentence membership variable $D^2_i$ and a document membership variable $D^3_i$. 
In this paper, we will overload $D^l$ ($l>1$) to indicate the higher-level group-memberships of
 lower-level-groups. For example if $g$ is the index of a level-2 group, then $D^3(g)$ is the level-3 group that covers all the datapoints under group $g$, 
i.e. $D^3(g)=D^3_i$ where $D^2_i=g$. Please see Fig~\ref{fig:group} for illustration.

Most topic models consider documents or sentences to be bags of words, and do not consider the sequential nature of the data. This can be avoided with the current representation, 
as sequential relations between the word-tokens can be encoded using the indices $\{D^1\}$ which takes integer values. Accordingly for each datapoint $i$ we can define sequential 
neighbors $prev(i)$ and $next(i)$. Even sequential ordering of the higher-level groups like sentences and documents can be captured by the variables $D^2$ and $D^3$ respectively. 
In case sequential ordering is irrelevant at any level (for example, ordering of documents is usually not relevant unless there are timestamps), the group membership variables at 
that level act as simple identifiers.

The groups at the different levels may be clustered in some applications, like multi-level clustering. For this, we associate a \emph{group cluster variable} with each 
group-index: $\{Z^1\},\{Z^2\},\dots,\{Z^L\}$. Again, we can overload $Z^l$ ($l>1$) to indicate the higher-level cluster memberships of lower-level groups. If $g$ is the index of 
a level-2 group, then $Z^3(g)$ is the level-3 cluster that covers all the datapoints under $g$, i.e. $Z^3(g)=Z^3_i$ where $D^2(i)=g$. This causes hierarchical clustering of the 
datapoints, specified by the tuple $\{Z^1_i,\dots,Z^L_i\}$.

\begin{figure}
 \centering
\includegraphics[width=3.5in,height=1.4in]{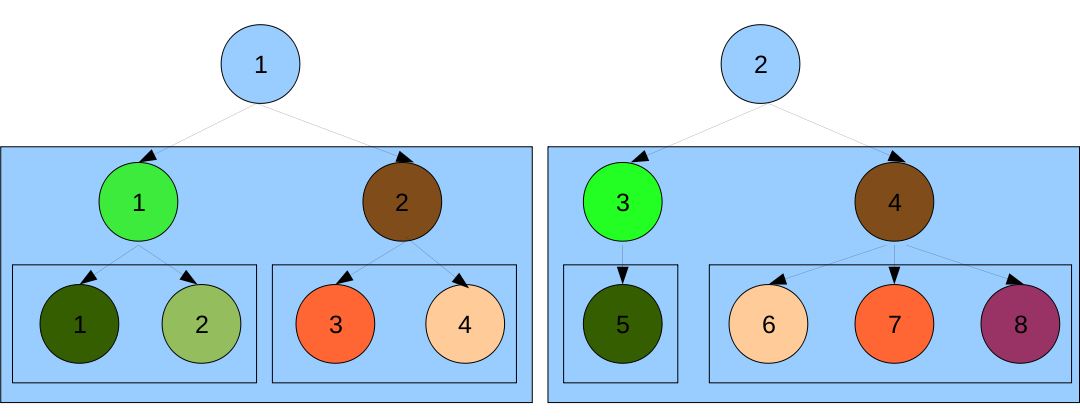} 
\caption{\small{Grouped data clustered at 2 levels ($l=1,2$). Colours indicate the clustering, like $Z^1_1=Z^1_5$, $Z^2(2)=Z^2(4)$ etc. 
Different colours used at the two levels. Note that $Z^1_3 \neq Z^1_6$, but $Z^2_3=Z^2_6$}}
\label{fig:clus}
\end{figure}

A Bayesian modelling involves mixture components and mixture distributions. We will consider $K$ mixture components (topics) $\phi_1,\dots,\phi_K$, where 
$K$ may not be known. Also, we need mixture distributions for each level- $\{\theta^1\},\{\theta^2\},\dots,\{\theta^L\}$. These are discrete distributions over index variables, 
that are cluster indices of the lower layer. Note that the cluster indices at level 1 are indices of the mixture components. At each level, the distributions may be specific to the 
group clusters, defined by the group cluster variables $Z$. For example, if the groups at level $l$ are clustered, the groups in cluster $k$ i.e. $\{j: Z^l(j)=k\}$ will have 
access to only the distribution $\theta^{l-1}_k$ at level $l-1$. The basic inference problem is to learn the cluster assignments $\{Z^l\}$, and estimate the mixture components 
$\phi$.

\section{Review of Existing Models}\label{sec:models}
In this section we make a short review of several well-known models using the above notation. The models can be classified based on the number of levels of grouping in the data 
that they consider.

\subsection{1-level models}
The simplest models are the 1-level mixture models, like GMM~\cite{bishop}. Here $L=1$, with $D^1_i=i$ and the datapoints are not grouped at all. 
There are $K$ mixture components $\{\phi\}$ which are Gaussian distributions, i.e. $\phi_k=\mathcal{N}(\mu_k,\Sigma_k)$. In general, the mixture components need not be Gaussian.
The mixture distribution $\theta^1$ is a $K$-dimensional multinomial. Each datapoint is assigned to a mixture component $Z^1_i$, which defines a clustering of the datapoints. This
 assignment is IID as $Z^1_i \sim \theta^1$ and sequential structure of the datapoints is not considered.  

In GMM, the number of mixture components $K$ is fixed and known. A non-parametric model with $L=1$ is the Dirichlet Process Mixture Model (DP-MM), which considers infinitely many 
mixture components, though only a few of them are used for a finite number of datapoints. The mixture distribution $\theta^1$ is an infinite-dimensional multinomial, 
drawn from a stick-breaking distribution. The parameters of the mixture components are drawn from \emph{base distribution} $H$. 

A one-level nonparametric model which does consider the \emph{sequential structure of the data} is the HDP-HMM~\cite{hdphmm}. This model considers a set of $\theta^1$-distributions
 from which one may be chosen conditioned on the previous assignments of $Z^1$. The $Z^1$-assignment to each datapoint $i$ is done as $Z^1_i \sim \theta^1_{j}$ with 
$j=Z^1_{prev(i)}$, where $prev(i)$ is the predecessor of the current datapoint in the sequential order encoded by $\{D^1\}$, i.e. $prev(i)=i'$ where $D^1_i=D^1_{i'}+1$.

\subsection{2-level models}
Next, we move into two-level models, i.e. where $L=2$. This is the standard setting for document modelling, where the word-tokens are grouped into documents (one level of grouping).
 The document membership of the variables are encoded by $D^2$. The most standard model of this kind is the Latent Dirichlet Allocation (LDA)~\cite{lda} which considers $K$ 
mixture components (topics) $\{\phi\}$, where $K$ is fixed and known. Each mixture component $\phi_k$ is a multinomial distribution over the vocabulary of size $V$. Here the 
level-2 groups (documents) are not clustered, i.e. $Z^2$ is distinct for each document. Consequently, $\theta^2$ is not used here, and $\{\theta^1\}$ are group-specific. The 
$Z^1$-variables of the datapoints within any group $j$ are assigned as IID draws from $\theta^1_j$. Once again, no sequential structure is considered.
Note that the mixture components $\phi$ are shared by all groups.
\begin{eqnarray}
 \phi_k \sim Dir(\beta), k \in [1,K]; \theta^1_j \sim Dir(\alpha), j \in [1,M] \nonumber \\
 Z^1_i \sim \theta^1_{D^2_i},  Y_i \sim \phi_{Z^1_i}
\end{eqnarray}
A non-parametric generalization of LDA is the Hierarchical Dirichlet Process (HDP)~\cite{hdp}, which is also a 2-level extension of the DP-MM discussed above. Here, the number of 
components is not fixed or known, so the document-specific $\{\theta^1\}$-distributions are infinite-dimensional, and drawn from a Dirichlet Process/Stick-Breaking Process
 instead of finite-dimensional Dirichlet.

Another nonparametric 2-level model is the Nested Dirichlet Process (NDP), where the level-2 groups (documents) are clustered using $Z^2$, which are drawn according to a discrete 
distribution $\theta^2$. Each cluster induced by $Z^2$ uses its own $\theta^1$. However, unlike the previous models, here the mixture components themselves are specific to the 
clusters induced by $Z^2$.
\begin{eqnarray}
 \phi_k \in H \forall k; \theta^1_z \sim GEM(\kappa_1) \forall z; \theta^2 \sim GEM(\kappa_2) \nonumber \\
 Z^2(j) \sim \theta^2, j \in [1,M]; Z^1_i \sim \theta^1_{Z^2(D^2_i)}, Y_i \sim \phi_{Z^2(D^2_i),Z^1_i}
\end{eqnarray}

\begin{figure}
 \centering
\includegraphics[width=1.7in,height=1.1in]{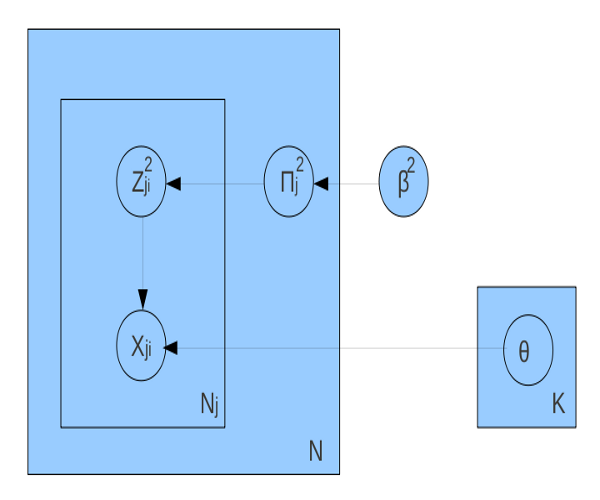}\includegraphics[width=1.7in,height=1.1in]{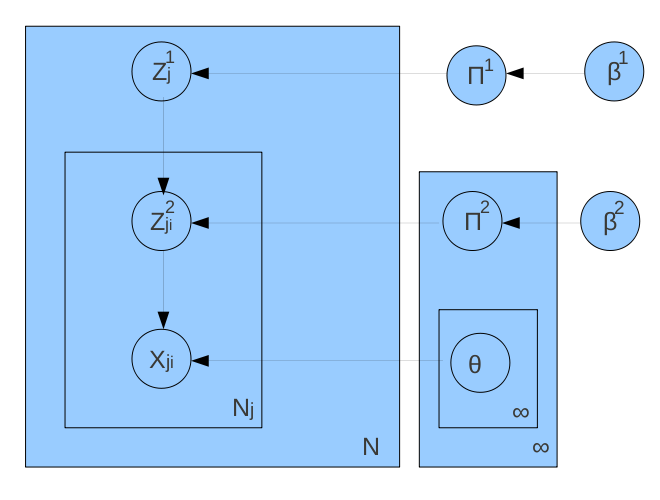}
\includegraphics[width=1.7in,height=1.1in]{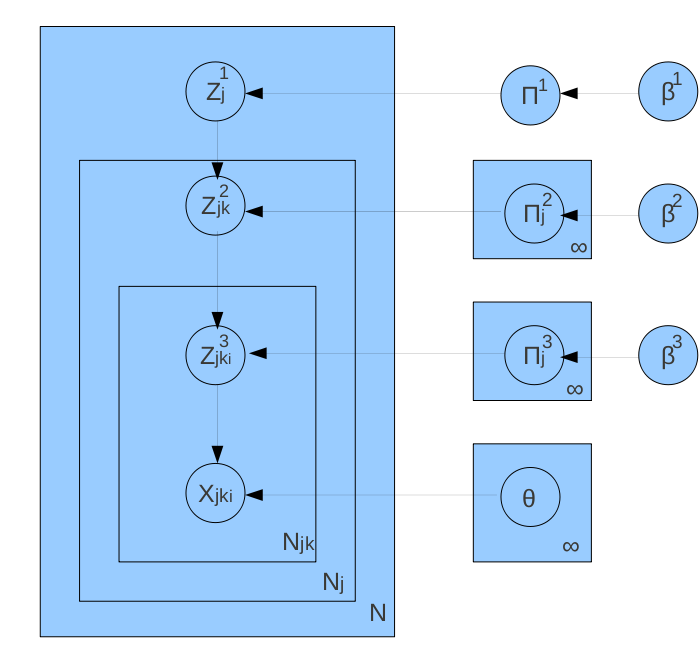}\includegraphics[width=1.7in,height=1.1in]{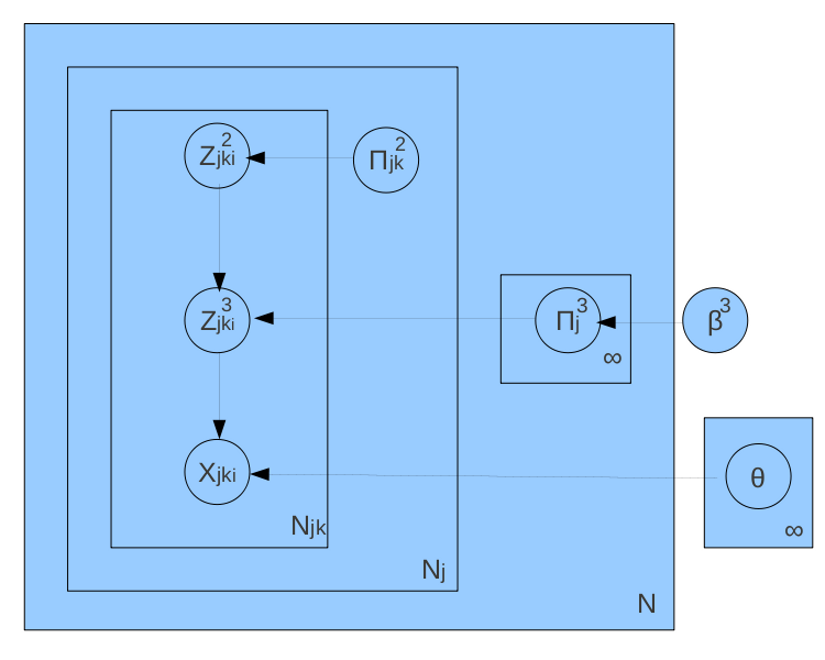}
\caption{\small{Above: HDP and NDP, Below: MLC-HDP and STM. The locations of the mixture components and distributions in the plate diagrams indicate the type of sharing 
(full/group-specific/cluster-specific)}}
\label{fig:2l}
\end{figure}

\subsection{3-level models}
Next, we look into some 3-level models. MLC-HDP~\cite{mlchdp} is an attempted compromise between HDP and NDP, where the groups are clustered (unlike HDP) but mixture components 
are not cluster-specific (unlike NDP), and moreover the data is grouped into 3 levels by observed group variables $\{(D^3,D^2,D^1)\}$. These groups can be clustered by random 
variables $\{Z^3\},\{Z^2\},\{Z^1\}$, which are drawn from discrete distributions $\theta^3$,$\{\theta^2\}$,$\{\theta^1\}$ respectively.

A three-level model that considers the sequential nature of the data is the Topic Segmentation Model (TSM)~\cite{strtm}. Here within each document the sentences are clustered 
using $\{Z^2\}$, but analogous to HDP-HMM the distributions $\theta^2$ are specific to values of $Z^2$. In particular, for any sentence $s$, $\theta^2$ is a distribution over two 
values: $Z^2(prev(s))$ and $Z^2(prev(s))+1$ (to induce linear clustering/segmentation). The $\{\theta^1\}$ are specific to the sentence-clusters. The documents 
themselves are not clustered, so $\theta^3$,$Z^3$ are not used. 

A somewhat unusual case is Subtle Topic Model (STM)~\cite{stm} which considers multiple document-specific distibutions over the mixture components, and distributions specific to 
sentences over this set of distributions. Here neither the documents nor the sentences are clustered. Effectively, only $\{\theta^1\}$-distributions are present, which are shared 
across sentences in the same document, but not across documents. However, the process of assigning $Z^1$-variables requires other sentence-specific variables in addition to 
$\{\theta^1\}$.


\section{DoS-classification of models}\label{sec:DoS}
In the above discussion, we have focussed on 3 major aspects- 1) Number of layers of grouping 2) the way in which the mixture components and mixture distributions are shared 
3) Whether sequential structure is considered or not at different layers. Based on these aspects, we propose a nomenclature for the models.

\subsection{DoS Concept}
As already discussed, in all the hierarchical Bayesian models, the mixture components $\{\phi\}$ and the mixture components $\{\theta^1\},\dots,\{\theta^L\}$ are shared among the 
different groups. We have seen three types of sharing 
\begin{enumerate}
 \item \emph{Full sharing (F)}: where components/distributions are shared by all the groups. For example, in HDP, MLC-HDP etc the mixture components are shared 
by all the level-2 groups.
 \item \emph{Group-specific sharing (G)}: where components/distributions are specific to groups, and not accessible outside the groups. For example, in HDP, STM etc the 
distributions $\theta^1$ are specific to the top-level groups (documents).
 \item \emph{Cluster-specific sharing (C)}: where the components/distributions are specific to clusters of groups, but not accessible outside the clusters. For example, in 
MLC-HDP each $\theta^1$-distribution is accessible to only one cluster of level-2 groups, and each $\theta^2$-distribution is accessible to only one cluster of level-3 groups. 
In all the models, the mixture components are specific to clusters of level-1 groups (as datapoints are clustered by the assignment of a mixture component through $Z^1$).
\end{enumerate}
Based on these notions we introduce Degree-of-sharing (DoS). For any given model, we first specify how the mixture components are shared at each of the levels- Full (F), 
group-specific (G) or cluster-specific (C), and we call this the DoS of $\{\phi\}$. The type of sharing at the different levels are hyphen-separated. Next, regarding the 
distributions $\{\theta^l\}$ at each level $l$, we specify how it is shared by the levels $(l+1)$ upwards, and we call this the DoS of $\{\theta^l\}$. 
Also, to indicate if sequential structure is considered at the different levels, we add $S$ to the levels where it is considered. Finally, to indicate how groups are clustered 
at different levels, we add $N$ to the levels where there is no clustering of groups, $P$ to the levels where the number of 
clusters is fixed, and $NP$ to the levels where the clustering is non-parametric. Note that this indicates the dimensionality of $\{\theta^l\}$- $P$ indicates that it is 
finite-dimensional, $NP$ indicates it is infinite-dimensional, and $N$ indicates it is not in use. 
 
By combining the DoS of $\{\phi\},\{\theta^1\},\dots,\theta^L$ in that order, we have the DoS-classification of the model. The DoS of the different variables are 
semicolon-separated. The number of components in any of these models is $(L+1)$, so that the DoS-classification of any model will have $(L+1)$ semicolon-separated parts. 
Also, the first part (corresponding to $\phi$) will consist of $L$ hyphen-separated letters, and the number of these letters will keep decreasing by one for each of the following 
parts (corresponding to $\theta^1$, $\theta^2$ etc), but followed by the letters specifying dimensionality and sequence strcuture. 

\subsection{Classification of Models}
Let us illustrate the concept of DoS-classification with a case study of all the models discussed in Section~\ref{sec:models}.

\textbf{Level-1} parametric models like GMM have mixture-components $\{\phi\}$ specific to clusters of datapoints, so that its DoS is $C$.
But the mixture distribution $\theta^1$ is fully shared by all the datapoints, so that its DoS is $F$. The number of clusters formed at level 1 (i.e. the dimensionality of 
$\theta^1$) is fixed (i.e. $P$) and sequential structure is not considered. Hence the DoS-classification of GMM is $C;F-P$. In case of DP-MM, $\theta^1$ is infinite-
dimensional (NP), i.e. DoS-classification is $C;F-NP$.

In case of HDP-HMM, the mixture-components $\{\phi\}$ are again specific to clusters of datapoints, so that its DoS is $C$. Here, the $\{\theta^1\}$ are non-parametric ($NP$), and
the sequential structure is also considered. So, the DoS-classification of HDP-HMM is $C;F-NP-S$. Note that here $\{\theta^1\}$ is a collection of distributions, from which one is 
chosen for each data-point $i$, depending on the assignment to $prev(i)$.

\textbf{Level-2 models:}
In HDP or LDA, the mixture-components are shared cluster-specific in level-1 and fully at level-2, so the DoS for $\phi$ is $C-F$. The  
$\{\theta^1\}$ are specific to level-2 groups, so the DoS for $\theta^1$ is $G$. Sequential structure is not considered at any level. In case of LDA the 
number of clusters of datapoints (level-1) is fixed ($P$), and for HDP it is $NP$. The level-2 groups are not clustered ($N$) in either model.
So we say the DoS-classification of LDA is $C-F;G-P;N$, and for HDP it is $C-F;G-NP;N$.
In case of NDP, the $\{\phi\}$ are cluster-specific at both levels, so its DoS is $C-C$. $\theta^1$ is specific to clusters of level-2 groups ($C$), and it is non-parametric 
($NP$). The $\theta^2$ are also non-parametric ($NP$). So, the DoS-classification is $C-C;C-NP;NP$.

\textbf{Level-3 models:}
In MLC-HDP, the mixture-components $\{\phi\}$ are cluster-specific in level-1, but fully at both levels 2 and 3, so that its DoS is $C-F-F$. The $\{\theta^1\}$ are 
specific to clusters of level-2 groups but fully shared by level-3 groups, and they are nonparametric, so the notation is $C-F-NP$. The $\{\theta^2\}$ are specific to clusters of 
level-3 groups and nonparametric, and finally $\theta^3$ is nonparametric. So the DoS-classification of MLC-HDP is $C-F-F;C-F-NP;C-NP;NP$.

For Topic-segmentation model, the topics $\{\phi\}$ are shared by all sentences and documents, i.e. its DoS is $C-F-F$. The $\{\theta^1\}$ are specific to clusters of sentences 
inside individual documents, i.e. the DoS is $C-G$, and they are of fixed dimension $(P)$. The $\{\theta^2\}$ used to cluster sentences is document-specific ($G$).
 The number of clusters (segments) of sentences to be formed is not fixed, and sequential structure is also taken into account, so the notation is $G-NP-S$. Finally, the 
documents themselves are not clustered ($N$), and the DoS-classification of TSM is $C-F-F;C-G-P;G-NP-S;N$.

Finally we come to Subtle Topic Model (STM), where the topics are shared by all sentences and documents, i.e. the DoS is $C-F-F$. The $\theta^1$ are shared by all sentences in a 
document but are specific to documents, and they are nonparametric, i.e. the notation is $F-G-NP$. The sentences and documents are not clustered, so the DoS-classification of 
STM is $C-F-F;F-G-NP;N;N$.

\section{Generalized Bayesian Model for Grouped Sequential Data}
Having discussed the DoS-classification of various existing models, it is clear that despite over a decade of research on topic models, there are several DoS-classifications for 
which there are no existing models. But instead of trying to point out those classifications individually and propose models following them, we now propose a generalized Bayesian 
Model for grouped sequential data. We will show that by specific settings of this model, it is possible to recover all the previously discussed models (or their close variants). 
Other models, not explored so far, can also be obtained from it.

\subsection{GBM-GSD}
We consider sequential data with $L$-levels of grouping, where the groups are sequential in every level (eg. in document modelling, we will consider the sentences within each 
document, and the documents themselves, are sequentially arranged). We consider that clustering happens at all levels, i.e. $\{\theta^1\},\dots,\{\theta^L\}$ all exist. To 
capture the sequential nature, we will assume that at every level (say $l$), there is a collection of distributions $\{\theta^l\}$ from which one can be chosen for each 
group, conditioned on the previous assignments (as considered in sHDP-HMM and TSM).
We also consider that all the distributions are infinite-dimensional (i.e. $NP$), i.e. neither the number of mixture components nor the number of clusters formed at each level is 
known in advance. We also consider that all the mixture components are accessible to all the level-2 groups, but introduce a binary random vector $B_i$ specific to each datapoint.
 This vector indicates which all mixture components are accessible to each datapoint. We will show that using this vector, we can make the mixture components group-specific or 
cluster-specific, and also capture other more intricate structures that would not be possible without it.
\begin{algorithm}[ht!]
\caption{Generalized Bayesian Model for Group Sequential Data (GBM-GSD)}
\begin{algorithmic}[1]
\scriptsize 
\STATE $\phi_k \sim H$, $\forall k$
\FOR{$g=1:G_L$}
\STATE $Z^L(g) \sim \theta^L|Z^L(1),\dots,Z^L(prev(g))$
\ENDFOR
\FOR{$l=L-1:2$}
\FOR{$g=1:G_l$}
\STATE $Z^l(g) \sim \theta^l_j|Z^l(1),\dots,Z^l(prev(g))$ where $j=Z^{l+1}(D^{l+1}(g))$
\ENDFOR
\ENDFOR
\FOR{$i=1:N$}
\STATE $B_i = f(B_1,\dots,B_{i-1},Z^1_1,\dots,Z^1_{i-1},Z^2,\dots,Z^L)$
\STATE $Z^1_i \sim B_i \circ \theta^1_{j}|Z^1_1,\dots,Z^1_{i-1}$ where $j=Z^2(D^2_i)$
\STATE $Y_i \sim \phi_{k}$ where $k=Z^1_i$
\ENDFOR
\end{algorithmic}
\end{algorithm}
The generative process hierarchically clusters the groups from top to bottom level. At every intermediate level $l$, it assigns $Z^l(g)$ to each group $g$ at level $l$. But for 
that it will have access to only those $\theta^l$-distributions, that are specific to the cluster $Z^{l+1}_g$ of group $g$ as a result of the clustering at level $(l+1)$. If 
group $g$ is part of group $m=D^{l+1}(g)$ at level $(l+1)$ then the $\theta^l$-distributions corresponding to $Z^{l+1}(m)$ must be used. Finally, at level 1, each datapoint $i$ is 
assigned a binary vector $B_i$ conditioned on the $B$-vectors corresponding to all previous datapoints. The distribution $\theta^1$ is convoluted with this vector $B_i$, so that 
a subset of the components are available to datapoint $i$.

\subsection{Recovery of Existing Models}
The level-1 models can be recovered easily. By setting $B_i$ as a vector of all $1$-s for all the datapoints, and by making $\theta^1$ conditioned only on $Z^1_{prev(g)}$ 
and GEM-distributed we get back HDP-HMM. In case $\theta^1$ is also independent of the previous assignments, we can have DP-MM, and if it is finite-dimensional it will GMM 
provided the base distribution $H$ is Gaussian.

When $L=2$, to recover HDP we need to define $\theta^L$ such that $Z^L(g)=g$ for all the groups $g$, so that groups are not clustered. Then we again set $B_i$ to be the vector 
of all $1$-s, and make $\{\theta^1\}$ independent of all previous assignments of $Z^1$. The $\{\theta^1\}$ should be drawn from a GEM. If the $\{\theta^1\}$ are 
finite-dimensional and drawn IID from a Dirichlet, and if all the $\{\phi_k\}$ are also drawn from a Dirichlet, then we have LDA.

NDP involves nonparametric clustering of level-2 groups without sequential ordering, so generation of $Z^L(g)$ should be independent of previous assignments, and $\theta^L$ 
should be drawn from a GEM. NDP also has the special characteristic that the different level-2 clusters do not share the same mixture components. This can be 
managed by setting $B_i$ through an appropriate function $f$, which will return a vector with $0$ for those mixture components that have been assigned in other level-2 clusters, 
i.e. $B_{ik}=0$ if $\exists j$  such that $Z^2_{D^1_j} \neq Z^2_{D^1_i}$ and $Z^1_j=k$.

When $L=3$, MLC-HDP can be recovered by removing the conditioning on previous assignments in the assignment of $Z^3$, $Z^2$ and $Z^1$, and by setting $B_i$ to be the 
vector of all $1$-s. The $\{\theta^l\}$ should be drawn from GEMs.
For TSM, $\theta^3$ should ensure that documents are not clustered, $B_i$ should be the vector of all $1$-s, and assignment of $Z^1$ should be independent of 
all previous assignments. Regarding $Z^2$, $\theta^2$ should ensure that for any sentence (level-2 group) $g$, $Z^2(g)$ should be either $Z^2(prev(g))$ or $Z^2(prev(g))+1$.

\section{News Transcript Segmentation}
We want to extend the generative framework for grouped sequential data (Algorithm 1) for modeling news transcripts. This data is hierarchical since there are broad news 
categories like politics, sports etc, under which there are individual stories or topics. In the Bayesian approach, we consider mixture components $\{\phi\}$ that correspond to 
these stories, and the broad categories are represented with distributions $\{\theta^1\}$ over these stories. As usual, each $\theta^1$-distribution is specific to a 
level-2 cluster (segment), and such clustering is induced by $\{\theta^2\}$, specific to the level-3 groups (the transcripts). The transcripts are not clustered.
The observed datapoints $Y_i$ are word-tokens, each represented as an integer (index of the word in the vocabulary). We define $prev(i)=i-1$ if ($i$, $i-1$) are in the same 
sentence, otherwise $prev(i)=-1$. Similarly, $next(i)$ is defined within sentences. Also, $prev$ and $next$ are defined for sentences. 
$Z^1_i$ indicates the news story (level-1) and $Z^2_i$ indicates the news category (level-2) that token $i$ is associated with. Each sentence is a level-2 group.

\subsection{LaDP model for news transcripts}
News transcripts were first modelled by Layered Dirichlet Process (LaDP)~\cite{ladp}. Several versions of this model was proposed, with different combinations of 
exchangeability properties at the different layers (which included MLC-HDP). Here, the level-2 groups were not sentences, but the word-tokens themselves.
In the most successful models, sequential structure was considered at both level-1 and level-2, i.e. the assignment of $Z^1_i$ and $Z^2_i$ are conditioned on $Z^1_{prev(i)}$ and 
$Z^2_{prev(i)}$ respectively. The clustering at both level-2 and level-1 are nonparametric. The DoS-classification of LaDP is $C-F-F;C-F-NP-S;F-NP-S;N$.

\subsection{Modeling Temporal Structure}\label{subsec:struc}
News transcripts have characteristic temporal features regarding assignments of $Z^2$ and $Z^1$ for which the GBP-GSD needs to be modified appropriately.
These features are discussed below. LaDP is insufficient for news transcripts, because it does not capture all of them.

\textbf{Number of Level-2 clusters (segments) are fixed and known}. In case of news transcripts from a particular source, it can be expected to have $K$ news categories in fixed 
order (say politics, national affairs, international affairs, business and sports, in that order).
\textbf{Segmentation} is the task of \emph{linear clustering} of words/sentences, i.e. each word/sentence $s$ can be assigned to either $Z^2_{prev(s)}$ or to $Z^2_{prev(s)}+1$. 
 In LaDP, each datapoint $i$ is assigned a value of $Z^1_i$ and $Z^2_i$ based on the assignments of $prev(i)$, and segmentation happens based on these assignments. But this does 
not guarantee the formation of $K$ segments. 
To overcome this issue, let it be known to model that the observed data sequence has $K$ level-2 segments. Then the sequence can be partitioned into $K$ parts of sizes 
$N_1,N_2,\dots,N_K$. These sizes may be modeled by a Dirichlet distribution where the parameters $\gamma_k$ signify the relative lengths/importance of the news categories.
\begin{scriptsize}
\begin{eqnarray}
 \{\frac{N_1}{N},\dots,\frac{N_K}{N}\} \sim Dir(\gamma_1,\dots,\gamma_K); Z^2_j= s \nonumber \\
\textbf{ where } \sum_{k=1}^{s-1}N_k < j \leq \sum_{k=1}^{s}N_k
\end{eqnarray}
\end{scriptsize}
In the GBS-GSD, the $\theta^2$ needs to be defined as a deterministic function, conditioned on $\{N_1,\dots,N_K\}$. 

\textbf{Topic Coherence} has been considered in various text segmentation paper like~\cite{topcoh}~\cite{ladp}. This is the property that within the same level-2 segment, 
successive datapoints are likely to be assigned to the same topic (mixture component). This can be easily modelled by the Markovian approach, i.e. 
\begin{scriptsize}
\begin{equation}
 Z^1_i \sim \rho\delta_{Z^1_{prev(i)}} + (1-\rho)(B_i\circ \theta^1_s) \textbf{where} s=Z^2_{D^2_i}
\end{equation}
\end{scriptsize}
This means that the $i$-th datapoint can be assigned the $Z^1$-value of its predecessor $pred(i)$ with probability $\rho$, or any value with probability $(1-\rho)$. The other 
available values are dictated by $B_i$, as discussed next. This is similar to the BE mixture model~\cite{ladp}.

\textbf{Level-2 segments do not share mixture components}, because each individual news story (topic) can come under only one news category. Also, 
\textbf{Topics do not repeat inside a Level-2 segment}. Inside a level-2 segment $s$, successive datapoints are expected to be assigned to the same mixture component 
due to temporal coherence. However, in news transcript, a news story will be told only once, which means that \emph{a particular component may be present only in a single 
chunk, and cannot reappear in non-contiguous parts of the segment.} For this purpose the generative process needs to be manipulated through $B_i$. 
Initially we set $B_i$ to be all 1s, and whenever a component $\phi_k$ is sampled for any point, we set $B_{ik}=0$ for all following points in the segment, so that 
$\phi_k$ cannot be sampled again. The generative process is as follows:
%
%
%
\begin{algorithm}[h!]
\caption{Generative Model for News Transcripts}
\begin{algorithmic}[1]
\scriptsize 
 \STATE $H_c \sim Dir(\beta)$ $\forall c$
 \STATE $c \sim U(K)$, $\phi_k \sim H_{c(k)}$ $\forall k$
 \STATE $\theta^1_s \sim GEM(\alpha)$ where $s \in [1,K]$
 \FOR{$g=1$ to $G^3$}
 \STATE $B_{gk}=1$ $\forall k$
 \STATE $\{\frac{N_{g1}}{N_g},\dots,\frac{N_{gK}}{N_g}\} \sim Dir(\gamma)$
 \ENDFOR
 \FOR{$j=1$ to $G^2$} 
 \STATE $Z^2_{j} = s$ based on $(N_{g1},\dots,N_{gK})$ where $g=D^3_j$
 \ENDFOR
 \FOR{$i=1:N$}
 \STATE if $Z^2_{D^2(i)}\neq Z^2_{D^2(prev(i))}$ set $\rho=0$
 \STATE $Z^1_i \sim \rho\delta_{Z^1_{prev(i)}} + (1-\rho)(B_g\circ \theta^1_s)$ where $s=Z^2_{D^2_i}$, $g=D^3_i$
 \STATE if $(Z^1_i \neq Z^1_{prev(i)})$ set $B_{gk}=0$ where $k=Z^1_i$, $g=D^3_i$
 \STATE $Y_i \sim mult(\phi_k)$ where $k=Z^1_i$
 \ENDFOR
\end{algorithmic}
\end{algorithm}

Here $G^3$ is the number of transcripts, and $G^2$ the number of sentences across all the transcripts. Clearly this model has 3 levels, and sequential structure is considered 
at level 2 (sentences) and at level 1 (word-tokens). Any topic $k$ belongs to a broad category $c(k)$ ($\in \{1,\dots,K\}$ uniformly at random), and corresponding to each category
 we have a base distribution $H_c$, which in turn are all drawn from a common base distribution $Dir(\beta)$. This helps to capture the fact that mixture components are specific 
to level-2 segments. The documents are not clustered, the sentences are clustered (segmented) with fixed number of segments, and 
the number of topics (word-clusters) is not fixed. The topics are shared across all transcripts, but are specific clusters of sentences, the $\theta^1$-distributions are 
specific to level-2 segments (clusters of sentences) but shared across transcripts, the $\theta^2$-distributions are transcripts-specific (parametrized by $\{N_g\}$) and 
$\theta^3$ are not used. So the DoS-classification for the generative model of news transcripts is $C-C-F;C-F-NP-S;G-P-S;N$.

\subsection{Inference Algorithm}
We now discuss inference for this model. We need an inference algorithm which ensures that $K$ segments are formed. We start with the joint distribution.
\begin{scriptsize}
 \begin{eqnarray}
 &p(Y,Z^1,Z^2,B,N,\Phi,\beta,\{\theta\},\{H\}) \propto \prod_{g=1}^{G^3}p(\{N_g\})\prod_{s=1}^K{p(\theta^1_s)} \nonumber \\ 
 &\times \prod_{c}{p(H_c|H)}\prod_{k}{p(\phi_k|H_{c(k)})} \prod_{j=2}^{G^2}{p(Z^2_j|Z^2_1,\dots,Z^2_{prev(j)},\{N\}})  \nonumber \\
 &\times \prod_{i=1}^N{p(Z^1_i,B_{next(i)}|Z^1_{prev(i)},B_i,Z^2_i,\{\theta^1\})}p(Y_i|Z^1_i,\Phi) \nonumber \\
\end{eqnarray}
\end{scriptsize}
We can collapse the variables $\{H\}$,$\{\Phi\}$, and $\{\theta^1\}$, and perform Gibbs Sampling. 
The key feature of this likelihood function is the presence of the $\{N_g\}$ variables. To handle these,
we introduce auxiliary variables $I_{g1},\dots, I_{g,K-1}$ which are the level-2 changepoints, i.e. the set of datapoints $\{i\}$ at which $Z^2_i \neq Z^2_{prev(i)}$.
Also note that $\{Z^2\}$, $\{I\}$ and $\{N\}$ are deterministically related. We introduce the $I$ variables to simplify the sampling.
We initialize the $Z^2$ variables by sampling a level-2 segmentation of the datapoints into $K$ segments. The $B$ and $Z^1$ variables are sampled accordingly. In each iteration 
of Gibbs Sampling, we consider the state-space of $I_{gs}$ as $I_{gs} \in \{I_{g,s-1},\dots,I_{g,s+1}\}$, i.e. the level-2 potential changepoints 
lying in between $I_{g,s-1}$ and $I_{g,s+1}$. The process is described in Algorithm~\ref{algo:gibgs}. Here, $B_{gs}=\{B_{set}\}$ where $set=\{i: D^3_i=g, Z^2_i=s\}$, i.e. the set 
of datapoints in transcript $g$ in segment $s$. (similarly $Z^2_{gs}$, $Z^1_{gs}, Y_{gs}$)
The major part in the Gibbs sampling is to sample the values ($\{B\}_s,\{Z^1\}_s$) for any segment $s$, conditioned on the remaining $B$ and $Z^1$ variables. This can be done 
using the Chinese Restaurant Process (CRP), where any component $k$ may be sampled for $Z^1_i$ (where datapoint $i$ is within segment $s$) proportional to the number of times it 
has been sampled, provided $B_{prev(i),k}=1$. The procedure is detailed in Algorithm 3, which is called Global Inference as it considers the overall structure of the transcript 
(as described in Sec\ref{subsec:struc}).
\begin{algorithm}[ht!]
\caption{Global Inference Algorithm by Blocked Gibbs Sampling (GI-BGS)}
\begin{algorithmic}[1]
\scriptsize
\FOR{transcript $g=1$ to $G^3$} 
\STATE Initialize $I_g$ with $(K-1)$ points by sampling from $Dir(\gamma)$; 
\STATE Set $\{Z^2\}$ according to $I$;
\STATE Initialize $\{B\}$,$\{Z^1\}$ variables;
\ENDFOR
\STATE Estimate components $\hat{\phi} \leftarrow E(\phi|Z,B,S,Y)$; 
\WHILE{Not Converged}
\FOR{transcript $g=1$ to $G^3$}
\FOR{segment $s=1:K$}
\STATE $I_{gs} \in \{succ(I_{g,s-1}),\dots,pred(I_{g,s+1})\} \propto p(Y_{gs}|B_{gs},Z^1_{gs},Z^2_{gs},\hat{\phi})$; 
\STATE Update $Z^2$ according to $I$
\STATE $(\{B\}_{gs},\{Z^1\}_{gs}) \propto p(\{B\}_{gs},\{Z^1\}_{gs}|\{B\}_{-gs},\{Z^1\}_{-gs},Z^2,Y,\hat{\phi})$;
\ENDFOR
\ENDFOR
\STATE Update components $\hat{\phi} \leftarrow E(\phi|B,Z^1,Z^2,Y)$;
\ENDWHILE
\end{algorithmic}\label{algo:gibgs}
\end{algorithm}

\section{Experiment on News Transcript Segmentation}
For news transcript segmentation, 
we used the news transcripts used by~\cite{ladp} for hierarchical segmentation. Here, each transcript has 4-5 news categories- politics, national affairs, international affairs, 
business, sports- in that fixed order. Overall, each transcript is about 5000 tokens long (after removal of stopwords and infrequent words), and has about 40 news stories, 
spread over the 5 categories. The task is to segment the transcript at two levels. At level 1, each segment should correspond to a single story, while at level 2, each segment 
should correspond  to a news category. The endpoints of the sentences are assumed to be known (these can be figured out based on pause durations in speech-to-text conversion), 
and are used to define level-2 groups. There are about 300-350 sentences per transcript.

In this dataset, the datapoints per sequence are too few in number to learn the level-1 mixture components (topics). Moreover, as already explained, each story occurs only once
in a transcript, thus reducing learnability. Hence, we considered 60 randomly chosen transcripts, and 
using initial segmentations of each sequence by the level-1 changepoints, 136 topics were learnt using HDP. These topics form our initial estimate of $\Phi$, using which 
we performed inference on individual sequences. The inference provides us with the $Z^2$ and $Z^1$ variables, based on which we can infer the segmentation at the two levels. 
We have gold standard segmentation available at both layers, and so we compute the segmentation errors $(S1,S2)$ at both layers. 
S1 and S2 can be computed by taking the average $P_k$-measure~\cite{ladp} for three different values of $k$, namely the maximum, minimum and average lengths of gold-standard 
segments (level-1 segments for S1 and level-2 segments for S2). 

We can look upon segmentation as a \emph{retrieval problem}, and define the \emph{Precision and Recall of level-2 segments (PR2 and RC2)}, and also for level-1 segments 
(PR1 and RC1). Let $i$ and $j$ be the starting and ending points of an inferred segment $s$, i.e. $Z^2_i=Z^2_{next(i)}=\dots=Z^2_j=s$, but $Z^2_{prev(i)}\neq s$ and 
$Z^2_{next(j)}\neq s$. Then, if there exists $(i0,j0,s0)$ such that $(i0,j0)$ defines a gold-standard segment $s0$ satisfying $|i-i0|<k$ and $|j-j0|<k$, then inferred segment 
$(i,j,s)$ is \emph{aligned} to gold standard segment $(i0,j0,s0)$. Precision, recall of a segmentation are defined as
\begin{tiny}
\begin{eqnarray}\label{eq:measures}
 \textbf{Precision}=\frac{\textbf{\#inferred segments aligned to a gold-standard segment}}{\textbf{\#inferred segments}} \nonumber \\
 \textbf{Recall}=\frac{\textbf{\#gold-standard segments aligned to an inferred segment}}{\textbf{\#gold-standard segments}} \nonumber
\end{eqnarray}
\end{tiny}
For level-2, the alignment threshold is set to 500, and at level-1 it is set to 10.
As a baseline, we use sticky HDP-HMM~\cite{hdphmm} and LaDP~\cite{ladp} at level-1, once again using the 136 HDP topics. For level-2, LaDP is the only baseline. We use the 
BE-BE-CE version, since that is the most successful according to~\cite{ladp}.

From the 60 news transcripts from which we learnt the 136 topics through HDP, we selected 3 (Trans1, Trans2, Trans3) to report the segmentation. Also, we selected another 3 
(Trans91, Trans92, Trans93) from outside the learning set, for which we used the same initial values of $\Phi$. The results are reported in Table 1. It is clear 
that in terms of all the measures we considered, GI-BGS outperformed both competitors at level-1. At level-2 also, GI-BGS is competitive on the three 
measures on all the transcripts except Trans1.

\begin{table}\label{tab:news}
\scriptsize
\begin{tabular}{| c || c | c | c || c | c | c || c | c | c |}
 \hline
\textbf{Data}  &     \multicolumn{3}{| c ||}{GI-BGS}       &     \multicolumn{3}{| c ||}{LaDP}         &     \multicolumn{3}{| c |}{sHDP-HMM}\\
\hline
                     &   PR1        &   RC1        &   S1        &   PR1        &   RC1        &   S1        &   PR1        &   RC1        &   S1  \\ 
\hline
Trans1               &\textbf{0.38} &\textbf{0.46} &\textbf{0.06}&   0.33       &\textbf{0.46} &  0.07       &   0.20       &   0.40       &  0.08 \\    
Trans2               &\textbf{0.33} &\textbf{0.37} &\textbf{0.10}&   0.27       &   0.34       &  0.11       &   0.18       &   0.34       &  0.12 \\
Trans3               &\textbf{0.26} &\textbf{0.41} &  0.09       &   0.25       &\textbf{0.41} &\textbf{0.08}&   0.13       &   0.32       &  0.11 \\  
Trans91              &\textbf{0.15} &\textbf{0.28} &  0.16       &   0.13       &\textbf{0.28} &\textbf{0.13}&   0.06       &   0.21       &  0.16 \\
Trans92              &\textbf{0.14} &\textbf{0.25} &  0.14       &   0.10       &   0.20       &  0.14       &   0.08       &   0.23       &  0.14 \\
Trans93              &\textbf{0.22} &\textbf{0.22} &\textbf{0.09}&   0.17       &   0.08       &  0.11       &   0.12       &   0.03       &  0.11 \\
\hline
\end{tabular}
\begin{tabular}{| c || c | c | c || c | c | c |}
 \hline
\textbf{Data}        &   PR2        &   RC2        &   S2        &   PR2        &   RC2        &   S2        \\   
 \hline
Trans1               &   0.20       &   0.20       &\textbf{0.08}&\textbf{0.33} &\textbf{0.40} &  0.11        \\
Trans2               &\textbf{0.80} &   1.00       &  0.04       &   0.71       &   1.00       &\textbf{0.01} \\
Trans3               &   1.00       &   1.00       &  0.13       &   1.00       &   1.00       &\textbf{0.04} \\
Trans91              &\textbf{0.60} &   0.60       &\textbf{0.06}&   0.50       &\textbf{0.80} &  0.07        \\
Trans92              &\textbf{0.60} &\textbf{0.60} &  0.05       &   0.20       &   0.40       &\textbf{0.04} \\
Trans93              &\textbf{0.60} &   0.75       &\textbf{0.04}&   0.50       &   0.75       &  0.06        \\
 \hline
\end{tabular}
\caption{\scriptsize{Above: Comparison of news transcript segmentation at level-1 by sticky HDP-HMM, LaDP and GI-BGS. Below: News transcript segmentation at level-2 by LaDP and 
GI-BGS.  Lower value of S1, S2 indicate better segmentation.}}
\end{table}

\section{Conclusions}
We carried out a study of various Bayesian models for hierarchically grouped data with emphasis on how they share mixture components and distributions 
among the groups. We also introduced the notion of Degree-of-Sharing (DoS) as a nomenclature for such models. We described a Generalized Bayesian model for this type of data, and 
showed how various existing models can be recovered from it. Next we used it to develop a new model for news transcripts, which has several peculiar temporal structures, and also 
provided an inference algorithm for hierarchical unsupervised segmentation of such transcripts. We showed that this model can outperform the existing LaDP model for this task. 
The DoS concept opens up possibilities to explore models with DoS-classifications that have not yet been considered, and the GBM-GSD  can be used to capture complex temporal 
structures in the data.

\bibliographystyle{named}
\bibliography{ijcai15}

\end{document}